\documentclass[times,twocolumn,final,authoryear]{elsarticle}

\usepackage{prletters}
\usepackage{framed,multirow}

\usepackage{amsmath}
\usepackage{amsthm}
\usepackage{latexsym}

\usepackage{url}
\usepackage{xcolor}
\usepackage{graphicx}
\usepackage{float}
\graphicspath{{./figures/}}
\DeclareGraphicsExtensions{.pdf,.png}
\usepackage{subcaption}
\usepackage{multicol}
\usepackage{natbib}

\definecolor{newcolor}{rgb}{.8,.349,.1}

\usepackage{rotating}
\usepackage{multirow}
\usepackage{yfonts}
\usepackage{comment}
\usepackage[draft]{todonotes}
\usepackage{xstring}
\newcommand{\note}[2]{%
    \IfEqCase{#1}{%
        {enable}{\par {\bfseries \color{blue} #2 \par}}%
        {disable}{}%
    }[\PackageError{note}{Undefined option to note: #1}{}]%
}%

\newcommand\inner[2]{\langle #1, #2 \rangle}
\newcommand\RotText[1]{\rotatebox{90}{\parbox{3.5cm}{\centering#1}}}
 
\journal{Pattern Recognition Letters}

\begin{document}

\begin{frontmatter}

\title{PDNet: Semantic Segmentation integrated with a Primal-Dual Network for Document binarization}

\author[1]{Kalyan Ram \snm{Ayyalasomayajula}\corref{cor1}} 
\cortext[cor1]{Corresponding author: 
  }
\ead{kalyan.ram@it.uu.se}
\author[1]{Filip \snm{Malmberg}}
\author[1]{Anders \snm{Brun}}

\address[1]{Division of Visual Information and Interaction, Dept. of Information Technology, Uppsala University, Uppsala, 751 05, Sweden.}

\received{1 May 2013}
\finalform{10 May 2013}
\accepted{13 May 2013}
\availableonline{15 May 2013}
\communicated{S. Sarkar}

\begin{abstract}
Binarization of digital documents is the task of classifying each pixel in an image of the document as belonging to the background (parchment/paper) or foreground (text/ink). Historical documents are often subjected to degradations, that make the task challenging. In the current work a deep neural network architecture is proposed that combines a fully convolutional network with an unrolled primal-dual network that can be trained end-to-end to achieve state of the art binarization on four out of seven datasets. Document binarization is formulated as an energy minimization problem. A fully convolutional neural network is trained for semantic segmentation of pixels that provides labeling cost associated with each pixel. This cost estimate is refined along the edges to compensate for any over or under estimation of the foreground class using a primal-dual approach. We provide necessary overview on proximal operator that facilitates theoretical underpinning required to train a primal-dual network using a gradient descent algorithm. Numerical instabilities encountered due to the recurrent nature of primal-dual approach are handled. We provide experimental results on document binarization competition dataset along with network changes and hyperparameter tuning required for stability and performance of the network. The network when pre-trained on synthetic dataset performs better as per the competition metrics. 
\end{abstract}


\end{frontmatter}


\section{Introduction}
\begin{center}
\begin{figure*}
\includegraphics[scale=0.4]{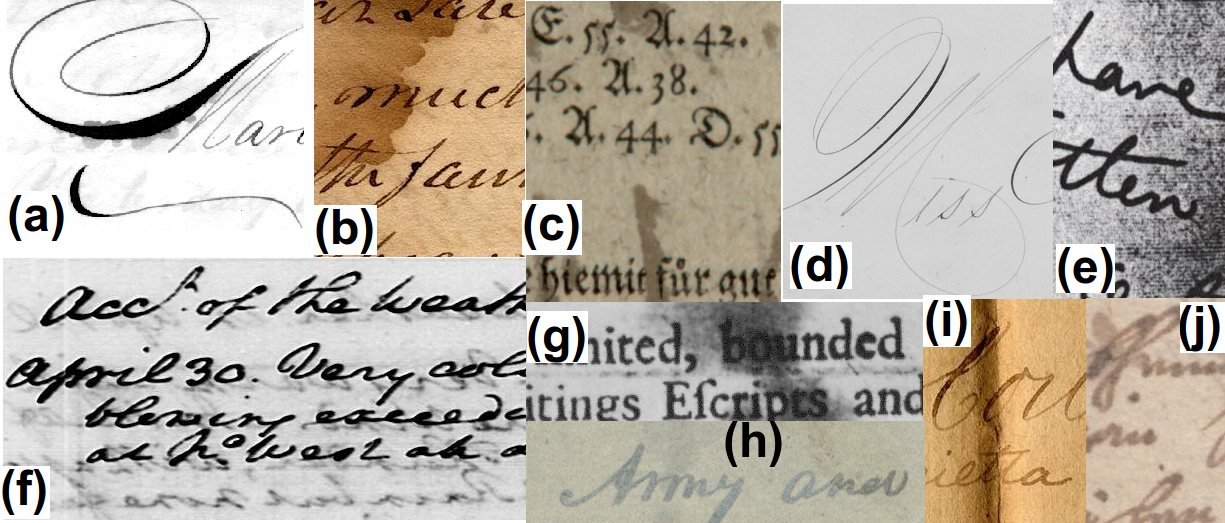}
\caption{Examples of typical image degradations from DIBCO dataset (a) smudging of text (b) staining of the parchment (c) textured background (d) uneven pen strokes (e) scanning artifacts (f) bleed through of ink from the other side of the document (g) blotting over text (h) feeble contrast between ink and parchment (i) artifacts from document aging (j) fading of text}
\label{fig001}
\end{figure*}
\end{center}

The process of binarizing digital documents deals with classifying each pixel as belonging to the background (parchment/paper) or foreground (text/ink) while preserving most of the relevant visual information in the image. Binarization is a common pre-processing step in most tasks performed on document images, such as word spotting and transcription where a high-quality and accurate binarization significantly simplifies the task at hand. In addition to the challenges due to uneven illumination and artifacts introduced by capturing devices, historical documents may have other degradations such as; bleed through; fading or paling of the ink in some areas; smudges, stains and blots covering the text; textured background and handwritten documents with heavy-feeble pen strokes for cursive or calligraphic effects to name a few. In general, this makes the task of document binarization very challenging as shown in Fig.\ref{fig001}. The task is often subjective with multiple acceptable outcomes encountered in corner cases such as considering ink blot as being part of foreground or background. The problem of document binarization garners interest in the field, which has led to the document image binarization content (DIBCO) \cite{Pratikakis11}, for automatic methods with minimum parameters to tune.

The task of document binarization borrows techniques from denoising, background removal, image segmentation and image in-painting, hence there exist several successful methods with individual strengths. The classical approaches have tried to separate the pixels into two classes using a single global threshold or series of finer local thresholds. The approach by \citet{Otsu79}, which tries to maximize the gray level separation between foreground (FG) and background (BG) classes by maximizing the inter class variance to separate the classes. However, local intensity variations and other artifacts introduced when creating a digital image have led to  the success of \emph{locally adaptive} techniques, such as the methods from \citet{Niblack86}, \citet{Sauvola}. The techniques discussed in all these classical methods are generic and applicable to any image in general. However, developing an approach specific to document images has been the trend in winning entries of DIBCO in the past. These methods seek to improve binarization through modeling properties of FG/BG in documents images specifically. \citet{Lu}, have for instance modeled background using polynomial smoothing followed by local thresholding on detected text strokes, \citet{Bar}, iteratively grow FG and BG within a $7 \times 7$ window.\\

The recent success of deep neural methods in vision related tasks has been broadly due to their ability to effectively encode the spatial dependencies and redundancy in an image. A fully convolutional neural network (FCNN) \citep{Shelhamer15}, is best suited for semantic labeling of pixels, which is the primary objective in segmentation. The crucial idea is to use skip connections to combine the coarse features from deep layers with fine features from shallow layers to improve the final segmentation. Training such models on text images, however, often result in loss of finer details along edges. Hence a post processing step such as a graph-cut \citep{GraphCut}, often improves the results as shown in our previous work \citep{Ayyalasomayajula17}. Here, we improve upon our previous work by incorporating the energy minimization step directly in the network to facilitate joint end-to-end training of both the semantic segmentation and energy minimization steps. To this end, we adopt a \emph{primal-dual update} ($\mathcal{PD}$Update scheme \citet{pdNet_Ochs}). See Fig.\ref{fig002}(a). This framework helps in training the \emph{unary cost} associated with pixel labeling, \emph{pairwise cost} \citep{BlakeMRF} associated with smoothness of neighboring pixels and the cost of overlooking an edge when merging regions into a single framework, resulting in an optimally segmented image. Our contributions in the present method can be summarized as:

\begin{figure*}
\begin{subfigure}{0.5\textwidth}
\centering
\includegraphics[scale=0.45]{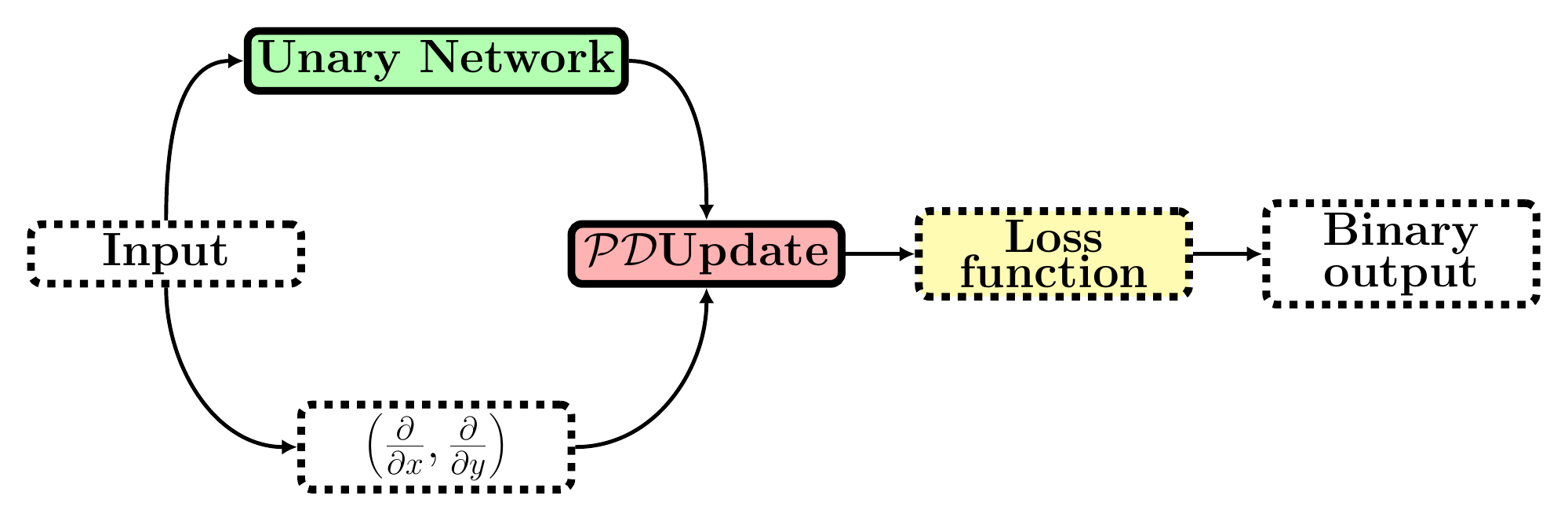}
\caption{}
\end{subfigure}
\begin{subfigure}{.5\textwidth}
\centering
\includegraphics[scale=0.1]{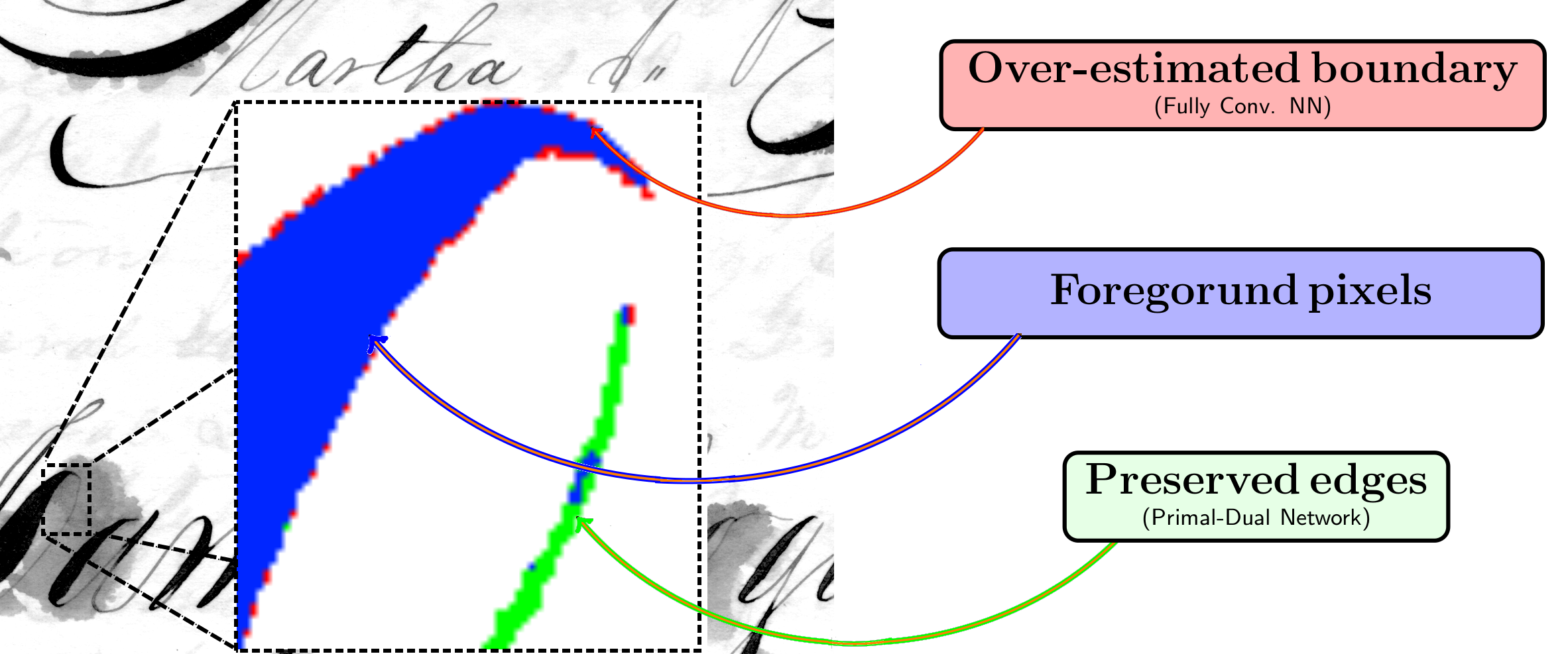}
\caption{}
\end{subfigure}
\caption{(a) Basic architecture of binarization network (PDNet) with unary network (ENet), primal-dual update ($\mathcal{PD}$Update), loss function and finite difference scheme based edge estimation blocks is depicted. Network modules are shown in solid lines and layers are shown as dashed lines. (b) This image summarizes the advantage of the PDNet over simple semantic segmentation. Upon zooming into a typical FG patch the pixels in blue are segmented by both ENet and PDNet. Pixels marked in red and green are over and under estimated by ENet. These are successfully delineated in PDNet due to $\mathcal{PD}$Update scheme. }
\label{fig002}
\end{figure*}

\begin{itemize}
 \item A stable framework that allows end-to-end training of an energy minimization function along with a semantic labeling network termed as Primal-Dual Net (PDNet).
 \item Improved segmentation output from a semantic labeling network that is lightweight in terms of trainable weights.
 \item A numerically stable, unrolled $\mathcal{PD}$Update scheme when formulating binarization as a \emph{total-variation} problem that can be extended to generic image based segmentation with multiple classes.
 \item Improved gradient propagation from $\mathcal{PD}$Update, using modified class weighting in loss function.
\end{itemize}

This paper is divided into five main sections describing the approach and an appendix that provides a summary of the necessary theory. An overview of document binarization with relevant background is covered in the introduction section. This is followed by a section reviewing some of the related work, highlighting some of the challenges that were overcome in our approach. An overview of the network architecture and functionality of its basic blocks are discussed in the methodology section, covering all the details on architectural changes made in building the network. This is followed by an experimental section that covers the results on DIBCO dataset for the architectural choices discussed previously. The article is concluded with contributions in the current work and possible directions for future research. All the necessary details on the theoretical framework is covered in an appendix towards the end.

\section{Related work}
 
The basic algorithm proposed in this paper draws motivations from other ideas that have used a high level loss function as an energy associated with binarization. This loss is then optimized by minimizing the said energy over the image; typical examples include the use of Markov random fields (MRF) for binarization \citep{Mishra} and intensity variation reduction using Laplacian kernel \citep{HoweConf}. These methods take both the global and the local aspects of the image into consideration to label the pixels. The former uses the Laplacian of the image to obtain invariance in BG intensity, followed by a graph-cut with suitable source-sink priors (seed points for FG-BG, respectively) and edge estimates required to build an image graph. Although the fundamental idea of using a defined loss as employed in Howe's method has been explored previously as separate methods, combining them into an energy function proved particularly effective. Further improvement of Howe's approach was proposed in our previous method \citep{Ayyalasomayajula14} by defining a \emph{3D binarization space} comprising of the intensity, horizontal and vertical derivatives at a pixel. A hierarchical clustering exploiting the inherent topology in this space, led to effective detection of seeds for the source and sink estimates and refinement of edges to improve the binarization result.\\

Before proceeding with the details in the method, we would like to motivate the reader towards an end-to-end trainable model as proposed in the current approach. The methods winning DIBCO competitions, 2009-2016 tune their parameters by training on the labeled data available from the previous competitions. Though this approach is common to all the supervised training methods in learning based approaches, these binarization methods fail to support their claim to generalizability through an exhaustive $k$-fold cross validation where the model is trained on all the labeled data from other competitions years and testing on data for the DIBCO competition year under consideration. We do compare our results with these state-of-the-art methods in Table.\ref{table02} however, it must be noted that these DIBCO competition winners are different for each year (\citet{ICDAR2009}, \citet{ICFHR2010}, \citet{ICDAR2011}, \citet{ICFHR2012}, \citet{ICDAR2013}, \citet{ICFHR2014}, \citet{ICFHR2016}).\\

As the current approach is based on a deep neural network we focus on FCNNs based approaches for further discussion. To our knowledge there are two methods based on FCNNs that provide an exhaustive cross-validation results on each of the DIBCO datasets for document binarization namely, \citet{Ayyalasomayajula17} and \citet{tensmeyer2017_binarization}. We used a post-processing based on graphcut in \citet{Ayyalasomayajula17}, to improve the segmentation output of a FCNN, however this approach is not end-to-end trainable. The proposed approach is an improvement over the former, as it is an end-to-end trained energy minimization approach formulated as a total-variation scheme on the FCNN output. As shown in Fig.\ref{fig002}(b) output from the later network compensates for the overestimated FG along the thick stroke boundaries and preserves the under estimated FG that is missed along the thin strokes by the former network. 

We conclude this sections by discussing some of the short comings in the other deep network proposed for document image binarization. As introduced in the previous section binarization can be solved through various formulations involving pixel classification. The approach that is relevant in all the FCNN based binarization methods is that of semantic segmentation. In the semantic segmentation literature \citep{semsegreview} integrating the context information surrounding a pixel showed an improved segmentation output. The proposed algorithm and \cite{tensmeyer2017_binarization} broadly fall into this category of refining the output from a FCNN. The former uses a total-variation framework and the later uses feature augmentation along with a loss function tailored for DIBCO metrics to meet this end. The performance of \cite{tensmeyer2017_binarization} depends two aspects. First, the binarization output is obtained from an ensemble of 5 networks trained on the DIBCO data instead of single trained network. Second, using feature augmentation such as image Laplacian \citep{HoweConf} or \emph{Relative Brightness feature}, which effect the binarization quality. \citet{ensemblebetter}, have shown that using an ensemble is always better than using a single trained network and \citet{hypercolumn} shown the advantage of feature augmentation on classification output of deep networks. Both these tricks though often used in training deep networks do not alter the fundamental segmentation output of the FCNN, instead improve the final output of segmentation after a suitable loss function. However, improving the fundamental output of FCNN is known to improve segmentation output \citep{crfrnn}. PDNet is generic network for semantic segmentation based on this idea of CRF unrolling \citep{crfrnn} that is aimed at improving the fundamental segmentation output from the FCNN, that achieves either state of the art or close to state of the art results on a single trained network instance instead of an ensemble.

\section{Methodology}

\begin{figure*}
\begin{subfigure}{0.5\textwidth}
\centering
\includegraphics[scale=0.4]{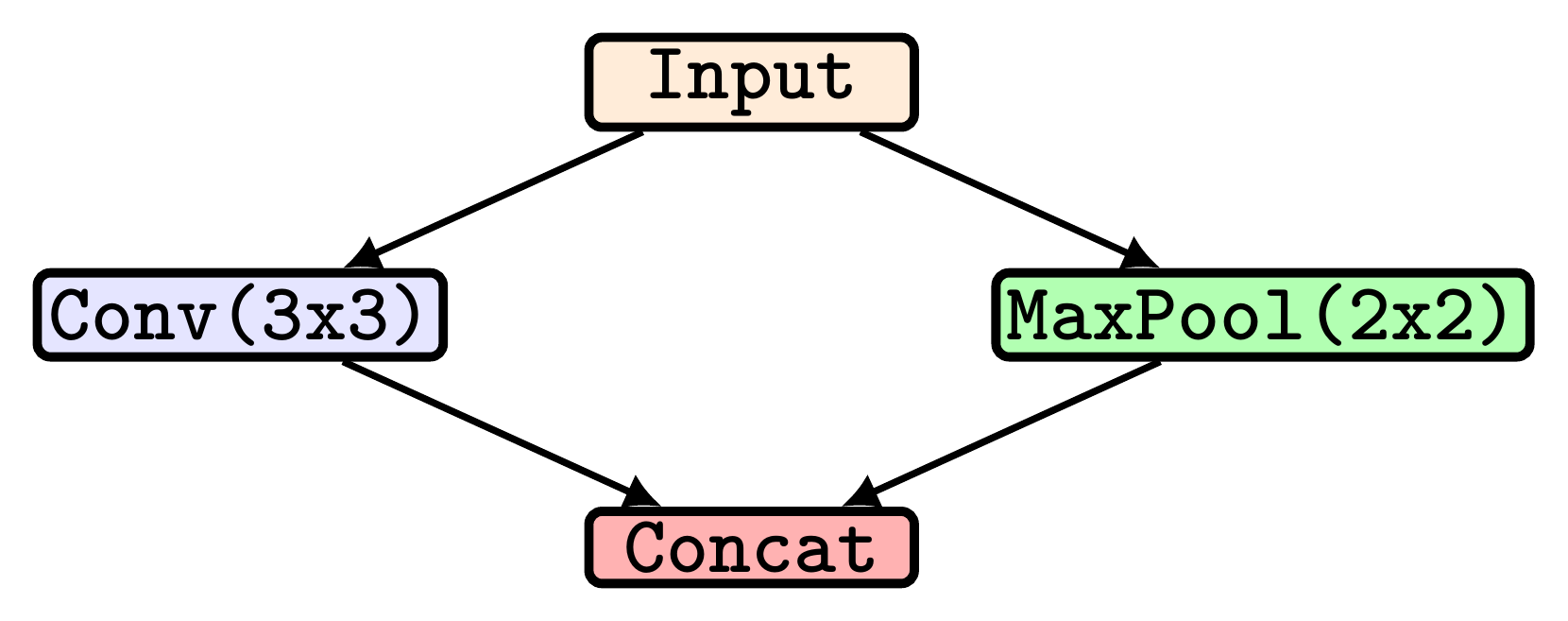}
\caption{}
\end{subfigure}
\begin{subfigure}{.5\textwidth}
\centering
\includegraphics[scale=0.4]{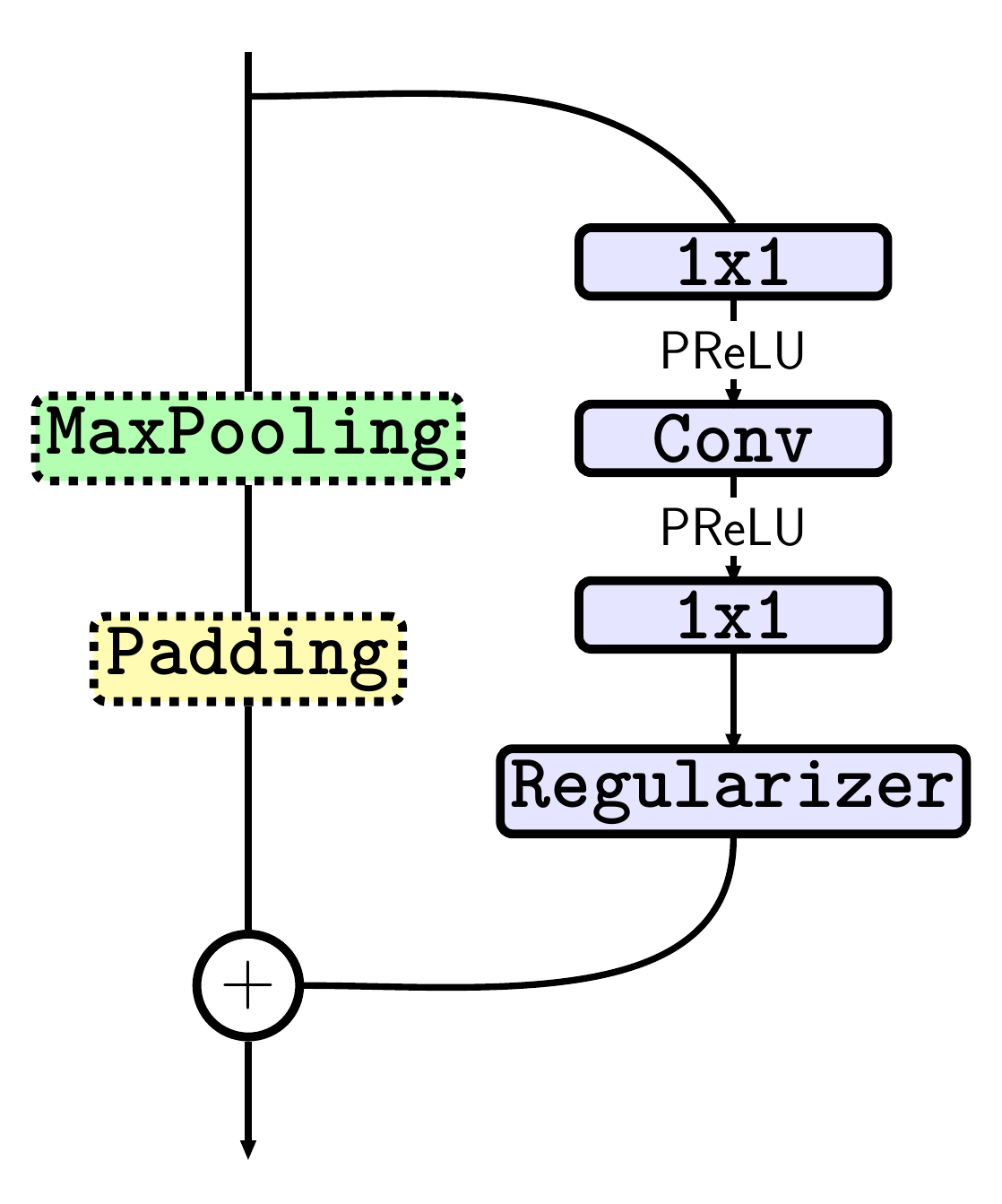}
\caption{}
\end{subfigure}
\caption{(a) ENet initial block. Convolution with kernel size $3\times3$, stride 2; MaxPooling is performed with non-overlapping 2 $\times$ 2 windows, there are 13 convolution filters, which sums up to 16 feature maps after concatenation. (b) ENet bottleneck module. conv is either a regular, dilated, or transposed convolution (also known as deconvolution) with 3$\times$3 filters, or a 5$\times$5 convolution decomposed into two asymmetric ones.}
\label{fig003}
\end{figure*}
\vspace{3mm}
The basic architecture of the proposed end-to-end binarization network PDNet is shown in Fig.\ref{fig002}(a). The network is built of three basic blocks:

\begin{itemize}
\item [--] {\bf \emph{Unary network}}: This is a semantic segmentation network that is capable of classifying each pixel in a given image into respective classes. Ideally such a network is quite capable of segmenting a given image by itself. Underlying such a classification is a cost associated with labeling each pixel as a particular class. We use the \emph{efficient neural network} (ENet) proposed by \cite{ENet_PaszkeCKC16}. The motivation for such an architecture is presented in the following section. However, as shown in our previous work \cite{Ayyalasomayajula14} and \cite{Ayyalasomayajula17}, text segmentation is sensitive to edge artifacts. Instead of using this network output directly we use the output prior to a typical softmax-like classification layer as the cost term for each pixel that can be further refined to improve the segmentation result.

\item [--] {\bf \emph{Primal-Dual Update}}: The design of this network is inspired from previous methods that made use of conditional random field (CRF) such as \cite{CRF_ChenPKMY14}, as a post processing layer in segmentation. This allows for a way to incorporate structural information among neighboring pixels in segmentation. However, we wanted to extend this idea further in text images, where the label propagation between neighbors should be encouraged but also restricted along the edges as shown in Fig.\ref{fig002}(b). Use of $\mathcal{PD}$Update scheme in tasks that involve total-variation formulation of energy function is already explored in depth super resolution \cite{pdNet_RieglerFRB16}, and multi-class labeling problem \cite{pdNet_Ochs}. We extend these ideas into a more stable architecture that permits end-to-end training within the intended theoretical framework of the underlying \emph{proximal operator}, eliminating the exploding gradient problem due to its recurrent structure.

\item [--] {\bf \emph{Loss function}}:
A typical document image has a lot of background pixels as opposed to written text, which naturally leads to class imbalance between the two classes in the training data. The loss function used in the network is a weighted Spatial Cross Entropy loss \cite{segNet_BadrinarayananK15}, often used to counter any class imbalance in the training samples. However, due to the redistribution of pixel labels resulting from the total-variation regularization as a part of the $\mathcal{PD}$Update these weights need to be re-adjusted. We propose an empirical approach to achieve this intended outcome.
\end{itemize}

\subsubsection{ENet architecture}
The ENet architecture is inspired from scene parsing CNNs based on probabilistic auto-encoders \cite{autoencoders_ICML2011Ngiam_399}, where two separate neural networks are combined as an encoder-decoder pair. The encoder is trained to classify an input through downsampling and the decoder is used to up sample the encoder output. The ENet architecture also tries to make use of the bottleneck module that was introduced in the ResNets \citep{resnets_HeZRS15}. A bottleneck structure consists of a main branch that is separated from an extension consisting of convolutions filters. These two branches are later merged using elementwise addition as shown in Fig.\ref{fig003}(b). The convolution layer \texttt{conv}, is either a regular, dilated or full convolution and if the bottleneck is downsampling then a max pooling layer is added to the main branch.\\

\begin{table}
\caption{ENet architecture for an example input of 512$\times$512, $C$ in the fullconv layer is the number of classes BN is the bottle neck layer indexed 1-5}
\label{table01}
\centering
\begin{tabular}[t]{|c| c c c| }
\hline
 & Name & Type & Output size \\ \hline
 & 		initial			&					& $16 \times 256 \times 256$\\ \hline
\multirow{12}{*}{\RotText{Encoder}} &		BN1.0			& downsampling 		& $64 \times 128 \times 128$\\
 &		$4\times$BN1.x	& 			 		& $64 \times 128 \times 128$\\ \cline{2-4}
 &		BN2.0			& downsampling		& $128 \times 64 \times 64$\\ 
 &		BN2.1			& 					& $128 \times 64 \times 64$\\ 
 &		BN2.2			& dilated 2			& $128 \times 64 \times 64$\\ 
 &		BN2.3			& asymmetric 5		& $128 \times 64 \times 64$\\ 
 &		BN2.4			& dilated 4			& $128 \times 64 \times 64$\\ 
 &		BN2.5			& 					& $128 \times 64 \times 64$\\ 
 &		BN2.6			& dilated 8			& $128 \times 64 \times 64$\\ 
 &		BN2.7			& asymmetric 5		& $128 \times 64 \times 64$\\ 
 &		BN2.8			& dilated 16		& $128 \times 64 \times 64$\\ \cline{2-4}
 & \multicolumn{3}{ c |}{\emph{Repeat section 2, without BN2.0}} \\ \hline
 \multirow{6}{*}{\RotText{\hspace{8mm}Decoder}} &		BN4.0			& upsampling		& $64 \times 128 \times 128$\\ 
 &		BN4.1			& 					& $64 \times 128 \times 128$\\ 
 &		BN4.2			& 					& $64 \times 128 \times 128$\\ \cline{2-4}
 &		BN5.0			& upsampling		& $16 \times 256 \times 256$\\ 
 &		BN5.1			& 					& $16 \times 256 \times 256$\\ \cline{2-4}
 & 		fullconv		&					& $C \times 512 \times 512$\\ \hline
  
\end{tabular}
\end{table}

We conclude the section by discussing a few key aspects of ENet  architecture as shown in Table.\ref{table01}. The encoder constitutes of the bottleneck sections 1-3 and sections 4,5 are part of the decoder. (indicated by BN{\bfseries{\emph{s.l}}}; {\bfseries{\emph{s}}} for section and {\bfseries{\emph{l}}} for layer within the section). ENet has some important architectural details that improve the speed of training keeping the parameters quite low. The projection layers do not include bias terms, to reduce the number of kernel calls and overall memory operations without effecting the accuracy. The network architecture heavily reduces the input size in the first two blocks allowing for small feature maps to be retained for further processing. This is because visual information can be highly compressed due to its inherent spatial redundancy. ENet opts for a large encoder with smaller decoder as opposed to a more symmetric design. This is motivated by the fact that encoder must be able to operate on smaller resolution data, reducing the role of the decoder to that of simple upsampling.\newline \\
ENet further exploits the redundancy in convolution weights by strategic replacement of $n \times n$ convolutions with two $n \times 1$ and $1 \times n$ convolutions filters as discussed in \cite{assymconv_JinDC14} and \cite{rethinkinception_SzegedyVISW15}. Strong downsampling of feature space needs to be compensated with equally adept up-sampling.  Convolutions layer were alternated with dilated convolutions to have a wide receptive field, while at the same time avoiding overly downsampling the input. Parametric rectifier linear Units (PReLU) \citep{prelu_HeZR015}, were used to learn the negative slope of non-linearities through an additional learnable parameter. Most of these aspects on receptive fields, non-linear activation functions and concerned limitation in text segmentation were raised in our previous work \citep{Ayyalasomayajula17}, making ENet architecture worth exploring for text binarization. 

\begin{figure*}
\centering
\includegraphics[scale=0.68]{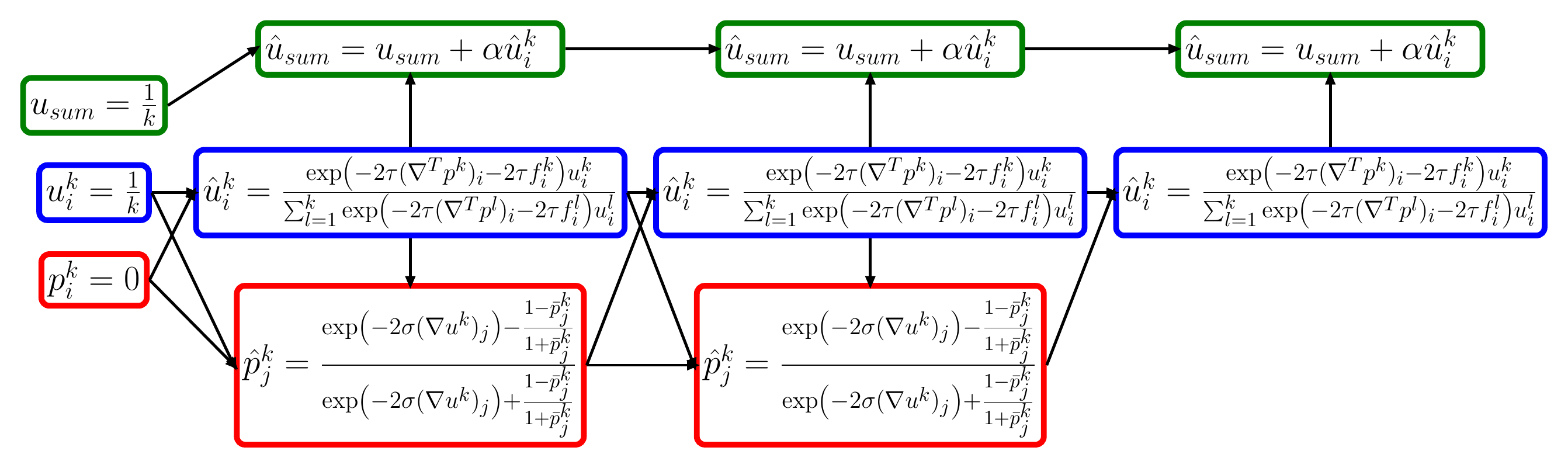}
\caption{$\mathcal{PD}$Update scheme unrolled thrice $\bar{p}_i^k$ is initialized to zeros, $\bar{u}_i^k, u_{sum}$ are initialized to $\frac{1}{k}$ the parameters $\tau, \sigma, \text{ and } \alpha$ are calculated in the network through gradient descent. The arrows indicate the dependencies of each block}
\label{fig004}
\end{figure*}

\subsubsection{ Primal-Dual Update scheme}
The \emph{primal-dual} update is built on three basic concepts 
\begin{itemize}
\item [--] \emph{Total variation} formulation for segmentation.
\item [--] \emph{Proximal operator} approach to decouple the underlying total variation and gradient operations.
\item [--] \emph{Bregman functions} based proximal operator smoothing to make proximal calculations differentiable.
\end{itemize}
as discussed in the Appendix section. The primal-dual formulation of the segmentation problem is given by
\begin{equation}
\min_{ u=(u_l)_{l=1}^k} \max_{ p=(p_l)_{l=1}^k} \left( \sum_{l=1}^{k} \Big( \inner{\nabla u_l}{p_l} + \inner{u_l}{f_l} \Big) \right) + \delta_U(u) - \delta_P(p)
\label{eq:primaldual}
\end{equation}
with the primal and dual updates given by 
\begin{equation}
\begin{split}
&\hat{u} = \Pi_{U} (\bar{u} - \tau \nabla^T p - \tau f ) \\
&\hat{p} = \Pi_{P} (\bar{p} + \sigma \nabla u )
\end{split}
\label{eq:proj_up}
\end{equation}
where are $u=(u_l)_{l=1}^k, p=(p_l)_{l=1}^k, f=(f_l)_{l=1}^k$ are the primal, dual and cost vectors for $1,\cdots,k$ classes respectively. $\delta_U(u), \delta_P(p)$ are the indicator functions for the primal and dual variables $u,p$ corresponding to the constraints sets $U,P$, respectively defined in Eqs.\ref{eq:unit_simplex},\ref{eq:inter_1balls}. The orthogonal projections on to $U,P$ are given by $\Pi_U, \Pi_P$, respectively . One approach to obtain a closed form representation for the projections in Eq.\ref{eq:proj_up} satisfying the constraints implicitly, is to use the \emph{Bregman proximity functions}. The updates for the $p,u$ are given in Eqs.\ref{dual_solution},\ref{primal_solution},  respectively. For the segmentation result to converge, the primal and dual updates need to be iterated over. PDNet has primal-dual updates unrolled over five times. A $\mathcal{PD}$Update with thrice such unrolled iteration is shown in Fig.\ref{fig004}, The $\bar{p}_i^k,\bar{u}_i^k, \text{ and } u_{sum}$ are initialized to $0,\frac{1}{k}$ and $\frac{1}{k}$, respectively. The overall output of the network can be interpreted as a perturbation of the unary cost using primal-dual updates to give a more controlled segmentation. The final segmentation is obtained by using a weighted cross entropy loss on the final $u_{sum}$.

\subsubsection{ Loss function}
A cross entropy criterion (CEC) combines the logistic-softmax over the class with classwise negative log likelihood criterion to obtain the final classification. A common problem in classification is imbalance in the samples over classes. This problem can be countered by adjusted the weights associating with each class, often estimated from the class histograms. However, due to the smoothing introduced by the $\mathcal{PD}$Update the weights for CEC loss estimated from the histogram alone overcompensate for the imbalance. We propose a power law over the histogram based weight calculation. The weighing used is inverse of square-root of class histograms. This power law is determined by the computing the training loss with $\mathcal{PD}$Update being part of the network iterated over various exponents as shown in Fig.\ref{fig005}(a).\\

The source code for the implementation of the PDNet used for binarization is made publicly available at \url{https://github.com/krayyalasomayajula/pdNet.git}

\subsection{Architectural modifications}
In the subsections below, we summarize the architectural changes from the basic network blocks.
\subsubsection{ E-Net architecture initial block changes}
When carrying out experiments on the DIBCO dataset we experimented with both color and gray scale images as well as using both color and gray channels. Although state of the art results were obtained using gray-scale images, we would like to highlight some extreme cases as shown in Fig.\ref{fig006}(a). When using color and gray channels the initial block was modified to include RGB and two gray channels with the max pooling applied to RGB channels alone; results for this case were similar to using RBG channels alone. When training the network for RGB and gray-scale alone the original ENet architecture was used without any changes.\\

\subsubsection{Clamped primal-dual updates}
\begin{figure*}
\begin{subfigure}{0.5\textwidth}
\centering
\includegraphics[scale=0.2]{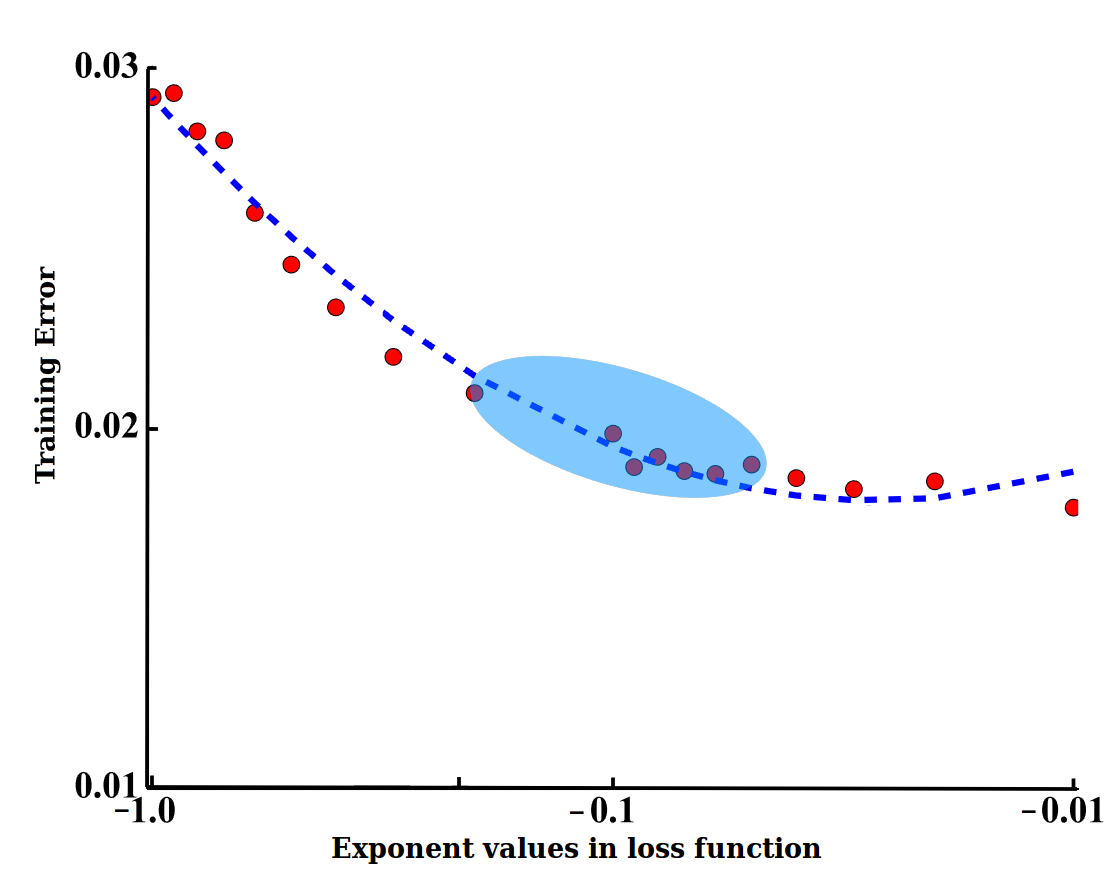}
\caption{}
\end{subfigure}
\begin{subfigure}{.5\textwidth}
\centering
\includegraphics[scale=0.2]{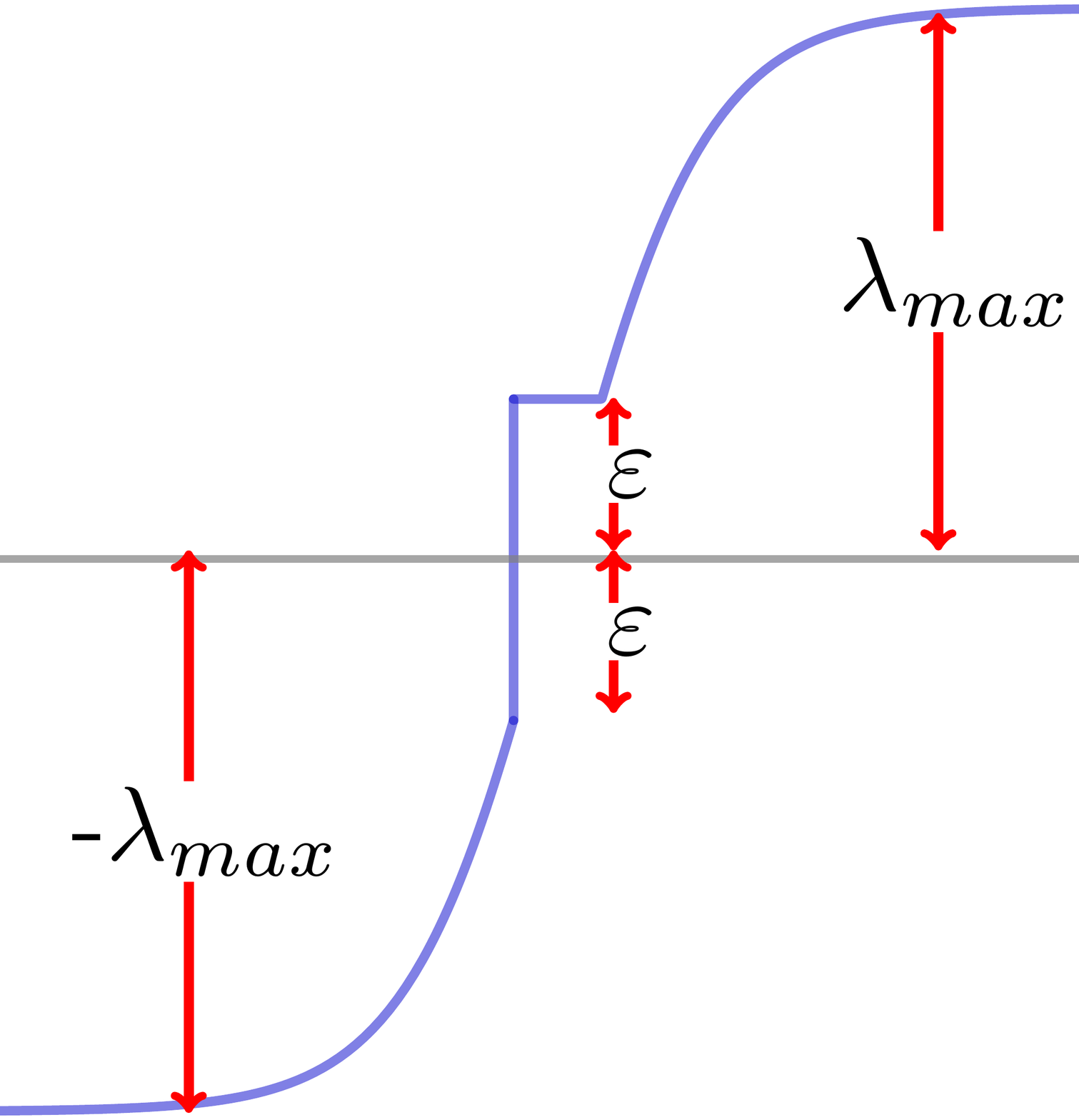}
\caption{}
\end{subfigure}
\caption{(a) Tuning of the exponent to be determined in the power law. The training error is plotted against exponent values form $[-1 ,-0.1] \cup [-0.1 ,-0.01]$ is steps of -0.1,-0.01 for the first and second intervals, respectively. The ellipse indicates range of suitable exponents that produce identical segmentation results. The exponent used in PDNet is -0.5. (b) The network learns permissible values for primal and dual variables $u_i,p_i$ in $\mathcal{PD}$Update by setting a large maximum value of $\lambda_{max}=10^{30}$ in positive and negative direction for them. At the same time values less than $|\varepsilon|$ are clamped to $10^{-8}$.}
\label{fig005}
\end{figure*}

The typical primal-dual updates when properly initialized are usually stable. However, when training a $\mathcal{PD}$Update scheme on image data in deep networks exploding gradient problem is commonly encountered. Implementations by \cite{pdNet_RieglerFRB16} and  \cite{pdNet_Ochs}, have dealt with this issue by gradient clipping to specific bounds during back-propagation. We resorted to another approach of clamping the values in primal and dual updates as shown in Fig.\ref{fig005}(b). This approach has two advantages:
\begin{itemize}
\item [--] The clamping as shown in Fig.\ref{fig005}(b) resets the pixel where instability was encountered to their initial values $0, \frac{1}{k}$ for $\bar{p}_i^k, \bar{u}_i^k$, respectively. The cost estimates for these pixels can be refined in further iterations, thus the resulting scheme is more faithful to the theoretical primal-dual approach.
\item [--] The gradients in this approach are not clipped during back-propagation thus leading to a faster training of the network making the loss converge within 10 epochs as opposed to 30 epochs in a network without clamping.
\end{itemize}

\subsubsection{Choice of labeling}
The convention often followed in semantic segmentation is to allow an \emph{unknown} class, to neglect objects such as background scenery or to include classes that need to be ignored. Generally, presence of such classes does not cause any change in the result, but do increase the training time for the PDNet. The  unknown class is usually labeled 1 and then other classes are labeled incrementally. We begin by relabeling the FG, BG and unknown classes to 1,2 and 3, respectively. Once the unary network is trained, the cost vector can be truncated to include just the FG and BG costs. Training PDNet on these costs leads to faster convergence as the size of $\bar{p}_i^k, \bar{u}_i^k$ and further vectors used in computations are reduced by 30\%.

\subsection{Tuning hyperparameters}
The binarization network has two hyperparameters to tune: The index to be used in the power law for the class weights used in the CEC loss function and the weights to be associated with the edges along the class boundaries. As mentioned previously the exponent in the power law for the weights was determined through an exhaustive search carried out on the training error for various exponents. The weights associated with the edges is a shared parameter over all the unrolled loops and multiplied with $\tau$ weight of each loop unroll. Training for both the edge weights and $\tau$ can cause instability. We instead initialized the edge weight to $1.0$ and trained on a small training-set to determine the convergence. This value was later used in the network. Although the edge weight determined by this approach is suboptimal, it is compensated for by learning $\tau$ values through back-propagation in the network over the complete training sets. This method of training for the edge weight produced a better validation loss, though this mode of pre-training is not critical for the network performance it can save time required for the over all training of the network.

\section{Experimental results}
\subsubsection{DIBCO dataset}
The experiments were conducted on the $DIBCO$ datasets by \cite{Pratikakis11}, for binarization competition from years 2009-2016 consisting of $76$ images in total. The images and ground truth images were augmented by applying an identical deformation field transformation to both. These augmented images were then converted into $128 \times 256$ pixels of cropped images with 25\% overlap horizontally and vertically to create more data for training. The cropped size $128 \times 256$ was selected based on the encoder requirement for the width and height to be equal to $2^n$ for some $n$ and to fit the training batches into memory in Torch7 framework \citep{torch}. The training set was picked by including the all the augmented images, except those from competition year under evaluation. The validation set was made by randomly picking 3000 original cropped images from the DIBCO datasets excluding the year under evaluation. The trained network was then used to produce binary output on the dataset of DIBCO competition year. The images were then stitched to the original size to compute the DIBCO evaluation metrics. Three main evaluation metrics for comparison are \emph{F-Measure}, \emph{Peak signal to noise ratio} (PSNR) and \emph{Distance Reciprocal Distortion Metric} (DRD), definitions of which can be found in \cite{Pratikakis11}, \cite{ICFHR2014}.

\begin{figure*}
\begin{subfigure}{0.5\textwidth}
\centering
\includegraphics[scale=0.23]{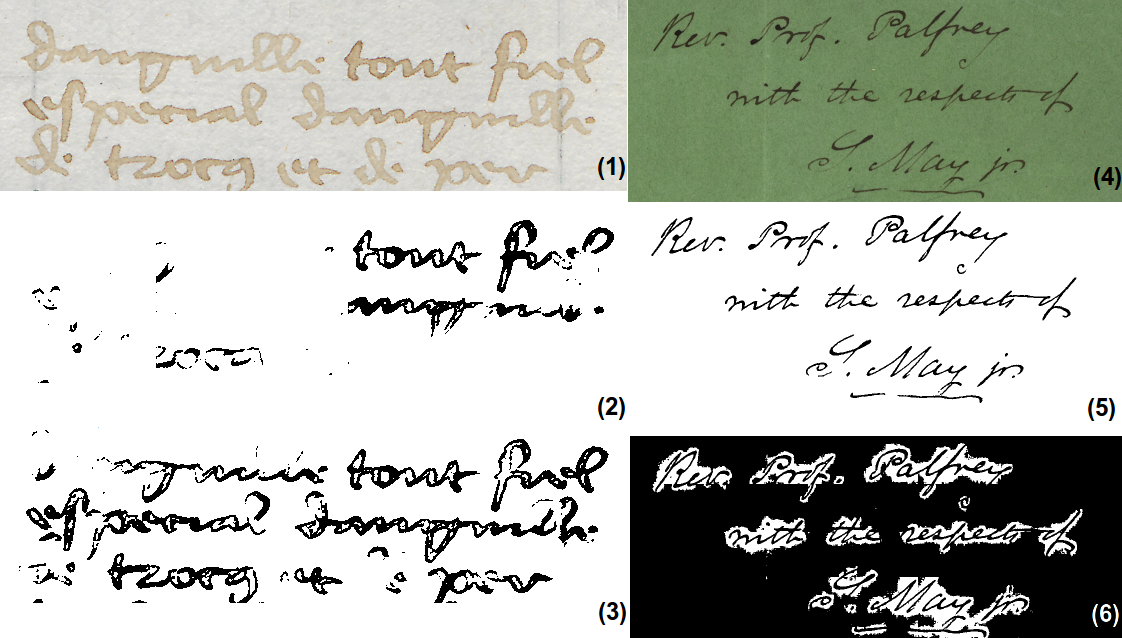}
\caption{}
\end{subfigure}
\begin{subfigure}{.5\textwidth}
\centering
\includegraphics[scale=0.37]{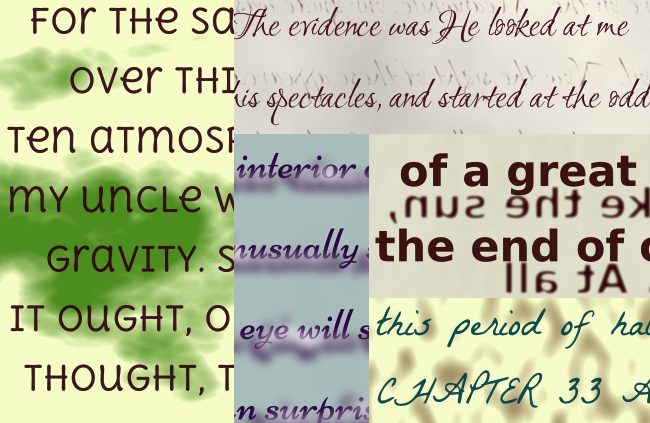}
\caption{}
\end{subfigure}
\caption{(a)(1) Document with low contrast between FG and BG (2) Binary output from network trained on gray scale images alone (3) Binary output using color input
(4) Document with a strong color BG (5) Binary output from network trained on gray scale images alone (6) Binary output using color input (b)Few samples from synthetic text data}
\label{fig006}
\end{figure*}

\begin{table*}[h]
\caption{Comparison of the results for F-Measure, PSNR, DRD for various methods. }
\label{table02}
\centering
\begin{tabular}[t]{|c|c|c|c|c|c||c|c|c|c|c||c|c|c|c|c|}
\hline
\multicolumn{1}{ |c| }{\multirow{2}{*}{Year}} &
\multicolumn{5}{ |c| }{FMeasure ($\uparrow$)} & \multicolumn{5}{|c| }{PSNR ($\uparrow$)} & \multicolumn{5}{ |c| }{DRD ($\downarrow$)} \\ \cline{2-16} &

\multicolumn{1}{ |c| }{PD$_g$} & PD$_c$ & GC & TM & DBC & PD$_g$ & PD$_c$ & GC & TM & DBC & PD$_g$ & PD$_c$ & GC & TM & DBC \\ \hline
\multicolumn{1}{ |c| }{2009} & {\bf 91.50} & 90.46 & 89.24 & 89.76 & 91.24 & {\bf 19.25} & 17.82 & 17.28 & 18.43 & 18.66 & {\bf 3.06} & 3.45  & 4.05 & 4.89 & - \\ \hline
\multicolumn{1}{ |c| }{2010} & 92.91 & 90.45 & 89.84 & {\bf 94.89} & 91.50 & 20.40 & 19.62 & 18.73 & {\bf 21.84} & 19.78 & 1.85 & 3.37  & 3.29 & {\bf 1.26} & - \\ \hline
\multicolumn{1}{ |c| }{2011} & 91.87 & 85.68 & 88.36 & {\bf 93.60} & 88.74 & 19.07 & 17.43 & 17.22 & {\bf 20.11} & 17.97 & 2.57 & 15.45 & 4.27 & {\bf 1.85} & 5.36 \\ \hline
\multicolumn{1}{ |c| }{2012} & {\bf 93.04} & 89.64 & 91.97 & 92.53 & 92.85 & 20.50 & 19.63 & 19.80 & 20.60 & {\bf 21.80} & 2.92 & 4.78  & 2.81 & {\bf 2.48} & 2.66 \\ \hline
\multicolumn{1}{ |c| }{2013} & {\bf 93.97} & 93.20 & 90.59 & 93.17 & 92.70 & {\bf 21.30} & 20.75 & 19.05 & 20.71 & 21.29 & {\bf 1.83} & 2.23  & 3.18 & 2.21 & 3.10 \\ \hline
\multicolumn{1}{ |c| }{2014} & 89.99 & 93.79 & 92.40 & 91.96 & {\bf 96.88} & 20.52 & 20.79 & 18.68 & 20.76 & {\bf 22.66} & 7.42 & 2.30  & 2.72 & 2.72 & {\bf 0.90} \\ \hline
\multicolumn{1}{ |c| }{2016} & {\bf 90.18} & 89.89 & 88.79 & 89.52 & 88.72 & {\bf 18.99} & 18.88 & 18.05 & 18.67 & 18.45 & {\bf 3.61} & 3.68  & 4.33 & 3.76 & 3.86  \\ \hline
\end{tabular}
\end{table*}

\subsubsection{Synthetic Dataset}
To have better weight initialization and prevent over fitting the network to data, the network was pre-trained on synthetic data. Documents resembling historical handwritten and printed material were generated synthetically. Various filters were applied to resemble background textures and degradations in the parchment. The text was generated using handwriting and machine printed fonts from the Google\texttrademark \citep{googlefonts}. Fig.\ref{fig006}(b) shows few cropped images from the synthetic dataset. The results from binarization on DIBCO dataset using the network pre-trained on the synthetic dataset are presented in Table-\ref{table02}.

\subsubsection{Training}
The network was trained in three stages: 
\begin{itemize}
\item [--] Pre-training unary network on synthetic dataset.
\item [--] Training unary network on DIBCO dataset.
\item [--] Combined training of unary and $\mathcal{PD}$Update scheme on DIBCO dataset.
\end{itemize}

The encoder of the unary network was pre-trained on patches of sizes $128\times256$ with batch-size 30 on the Synthetic dataset. The pre-trained encoder was then used to initialize the decoder weights and then the decoder was per-trained on the Synthetic dataset using the same patch size and batch size. This gives the pre-trained unary model. This model was then used to initialize the weights of the unary model that is to be trained on the augmented DIBCO dataset. The gradients were trained using \emph{ADAM} \citep{adam_KingmaB14}, with the  learning rate, weight decay, and momentum set to $5 \times 10^{-4}, 2 \times 10^{-4} \text{ and } 0.9$, respectively. The unary models were trained for 20 epochs and the best model with least loss was picked in further steps. When training the unary and $\mathcal{PD}$Update the learning rate was set to $5 \times 10^{-4}$ for the first 10 epochs and decayed to $2 \times 10^{-4}  \text{ and } 1 \times 10^{-4}$ between 11 - 15 and 16 - 20 epochs, respectively. The results on DIBCO dataset, with the proposed binarization network is summarized in Table.\ref{table02}.\\

PD$_g$, PD$_c$ are the outputs from primal-dual networks trained on grayscale and color images, respectively. GC is the result obtained from a graph-cut based approach as discussed in our previous work \cite{Ayyalasomayajula17}. It takes the output from the unary model as seed points along with the output from the classification layer acting as costs for pixels and Canny edges as boundaries estimates for segmentation. This approach requires three parameters a) cost associated with pixels labeling as obtained from the unary network, 2) weight associated to edges and Canny threshold to estimates the boundaries. It then employees an external graph-cut method \citep{GraphCut}, to obtain the final segmentation. In contrast to this approach the current methods does all these parameter estimations and the energy minimization in end-to-end manner in a single framework. TM are the results from another CNN based approach developed by \cite{tensmeyer2017_binarization}, which augments the segmentation result with \emph{relative darkness feature} \cite{relativedarkness_WuRAN15}, to aid in binarization. DBC are results from winning entries in DIBCO competitions using various classical approaches in  as given in \cite{ICFHR2014}, \cite{ICFHR2016}.

\section{Conclusion}
In the current work we have extended our previous approach of using graph-cuts as post-processing to a FCNN, to obtain binarized documents images. In this paper, we propose an architecture that combines the use of an energy minimization function and FCNN based feature learning. The energy minimization framework is flexible in imposing constraints on the desired segmentation. This primal-dual network imposes a total variation based energy minimization, with an unrolled primal-dual update scheme in a gradient descent trainable CNN architecture. The unary CNN, on the other hand, learns pixel labeling costs and the combined network learns all the associated parameters within a single end-to-end trainable framework.
\vspace{3mm}\\
The numerical instability caused by the recurrent nature of the primal-dual steps was solved by using a clamping function on the primal and dual updates within the $\mathcal{PD}$Update block. Propagation of gradients from the $\mathcal{PD}$Update was further facilitated by employing a weighted cross entropy loss adjusted by a power law. The final primal-dual architecture improves the binarization results, compared to the FCNN output alone, even when combined post-processing such as graph-cut or feature augmentation. The binarization achieves state of the art results on four out of seven DIBCO datasets, with comparable to best result on the rest. Further investigation in dropout layers in the unary model and architectural variations in $\mathcal{PD}$Update are planned for future work. The results from using a trained network to other historical document images and cases of transfer learning are of practical interest. Using more sophisticated unary and primal-dual schemes could also yield an even better result on document binarization.

\section*{Acknowledgments}
This project is a part of q2b, From quill to bytes, an initiative sponsored by the Swedish Research Council "Vetenskapsr\r{a}det D.Nr 2012-5743) and Riksbankens Jubileumsfond (R.Nr NHS14-2068:1) and Uppsala university. The authors would like to thank Tomas Wilkinson of Dept. of Information Tech., Uppsala University for discussions on debugging and performance optimization in Torch framework.

\bibliographystyle{model2-names}
\bibliography{refs}

\section*{Appendix}
\label{sec:appendix}
{\bf \emph{Total variation formulation:}} Formulating image segmentation as a saddle point problem and applying proximal operator to the primal-dual variables is a well studied problem in fixed point analysis of convex functions \citep{Chambolle2011}. An overview of theory is provided here to make the discussion comprehensive and to provide intuition for $\mathcal{PD}$Update discussed in the method. We consider segmentation of image into $k$-pairwise  disjoint regions as a total variation on the segmentation image as: 
\begin{equation}
\min_{(R_l)_{l=1}^k, (c_l)_{l=1}^k} \frac{1}{2} \sum_{l=1}^{k} Per(R_l,\Omega) + \frac{\lambda}{2} \sum_{l=1}^{k} \int_{R_l} |g(x)-c_l|^2 dx,
\label{eq:tv_seg}
\end{equation}
where $Per(R_l,\Omega)$ is the perimeter of the region $R_l$ in a domain $\Omega$, $g: \Omega \rightarrow \mathbb{R}$ is the input image, $c_l \in \mathbb{R}$ are the optimal mean values and the regions $(R_l)_{l=1}^k$ form a partition of $\Omega$ that is, $R_l \cap R_m = \emptyset, l \ne m$ and $\cup_{l=1}^k R_l = \Omega$, $\lambda$ is a regularization weight. This is an optimization problem between the data fitting term $|g(x)-c_l|$ and the length term
 $Per(R_l,\Omega)$ where the ideal mean values of the region $c_l = \frac{\int_{R_l} g(x)}{|R_l|}$ are unknown \emph{a-priori} as they depend on the partition we seek.\\
 
By introducing a labeling function $u = (u_l)_{l=0}^k: \Omega \rightarrow \mathbb{R}$ and viewing the data fitting term as a weighing function $f_l = \frac{\lambda}{2}|g(x)-c_l|$ Eq(\ref{eq:tv_seg}) can be generalized as:
\begin{equation}
\begin{split}
&\min_{ u=(u_l)_{l=1}^k} J(u) + \sum_{l=1}^{k} \int_{\Omega} u_l f_l dx, \\ s.t \hspace{2mm} &\sum_{l=1}^{k} u_l(x)=1, u_l \ge 0, \forall x \in \Omega,
\end{split}
\label{eq:relaxation_seg}
\end{equation}
where $J(u)$ is the relaxation term. \\

{\bf \emph{Proximal operator:}}
This basic structure in Eq.\ref{eq:relaxation_seg} is of the form 
\begin{equation}
\min_{x \in X} F(Kx) + G(x),
\label{gen_primal}
\end{equation}
involving a linear map $K: X \rightarrow Y$ with the induced norm $||K|| = \max\left\{ ||Kx||: x \in X \text{ with } ||x|| \le 1 \right \}$ where $X,Y$ are the primal and dual spaces, respectively. The corresponding dual formulation of this equation is a generic \emph{saddle-point} problem 
\begin{equation}
\min_{x \in X} \max_{y \in Y} \inner{Kx}{y} +G(x) - F^*(y),
\label{gen_saddle}
\end{equation}
where $\inner{Kx}{y}$ is the inner product induced by the vector space $Y$ and $F^*$ is the \emph{convex conjugate} of $F$.\\

The advantage of such an approach is discussed further down with the segmentation example, for now we can observe that this structure readily presents a computationally tractable algorithm. The dual variable $y \in Y$ acts like bounded slack variables introduced to ease the solution in the resulting dual space $Y$. Introducing the dual variable $y$ relieves the initial composition of $F(Kx)$ making computations involving $\inner{Kx}{y}$ independent of $F^*(y)$. As per the structure of segmentation problem, $F$ is typically an indicator function for capturing the constraints on $x$, which translates to $F^*$ being an indicator function for capturing the constraints on its dual variable $y$. Since the primal space has $||x|| \le 1$ if the dual variable is bounded, which is most often the case then iterating repeatedly between the two variables should converge to a solution.\\

The solution takes a form involving the \emph{proximal operator} or gradient of the functions $F,G$ depending on them being convex or convex as well as differentiable, respectively. The basic idea behind a proximal operator is a generalization of projection on to a vector space. This makes it an ideal operator that can be used in a gradient descent algorithm where the iteration involves taking a suitable step towards the solution along the gradient direction. But since the function need not be differentiable the gradient need not necessarily exist and hence the question of uniqueness along the gradient direction does not arise. This results in a set of permissible vectors that though not strictly a gradient can act as one at a given point $x$, such a set of permissible vectors is called \emph{subgradient}. The set $\partial F$ is the subgradient it is also the set of $underestimators$ of $F$ at $x$. A closely related set is the \emph{resolvent operator} with the property
\begin{equation}
x = (I + \tau \partial F )^{-1} (y) = \arg \min_{x} \left\{ \frac{||x-y||^2}{2\tau} +F(x) \right\}.
\label{resolvent}
\end{equation}
It can be interpreted as the closest point $x$ in the set under consideration to the given point $y$ under an associated error $F(x)$. The primal-dual formulation allows for an algorithm that iterates between the primal and dual variables $x,y$, respectively in this case leading to convergence according to the $forward-backward$ algorithm
\begin{equation}
\begin{split}
&y^{n+1}=prox_{x}(y^n) = (I + \sigma \partial F^* )^{-1} (y^n + \sigma K \bar{x}^n); \\ 
&x^{n+1}=prox_{y}(x^n) = (I + \tau \partial G )^{-1} (x^n - \tau K^* {y}^{n+1});\\
& \bar{x}^{n+1} = x^{n+1} + \theta (x^{n+1} - x^{n}),
\end{split}
\label{eq:proximal_alog}
\end{equation}
where $\tau$, $\sigma$ are step lengths along dual and primal subgradients and $\theta$ is the relaxation parameter in iterating the primal variable.\\

Considering the image discretised over the Cartesian grid of size $M \times N$ as $\left\{ (ih,jh) : 1 \le i \le M, 1 \le j \le N \right\}$, where $h$ is the size spacing and $(i,j)$ the indices in discrete notation. $X$ is a vector space in $\mathbb{R}^{MN}$ equipped with standard inner product $\inner{u}{v}$ for $u,v \in X$. The gradient is defined as $\nabla: X \rightarrow Y$, $\nabla u = (\frac{u_{i+1,j}-u_{i,j}}{h},\frac{u_{i,j+1}-u_{i,j}}{h})$ with $Y=X \times X$ equipped with the inner product defined as,
\begin{equation*}
\inner{p}{q}_Y = \sum_{i,j} p_{i,j}^1 q_{i,j}^1 + p_{i,j}^2 q_{i,j}^2, p=(p^1,p^2), q=(q^1,q^2) \in Y.
\end{equation*}

Applying the above framework to Eq.\ref{eq:relaxation_seg} with $J(u) = \frac{1}{2} \sum_{l=1}^{k} \int_{\Omega}|\nabla u_l|$, we have
\begin{equation}
\min_{ u=(u_l)_{l=1}^k} \frac{1}{2} \sum_{l=1}^{k} \left( \int_{\Omega}|\nabla u_l| + \inner{u_l}{f_l} \right) + \delta_U(u),
\label{eq:relaxation_primal}
\end{equation}
where $G(u) = \delta_U(u)$ is the indicator function for the unit simplex,
\begin{equation}
U = \left\{ u \in X^k : \sum_{l=1}^{k} u_l(x)=1, u_l \ge 0 \right\}.
\label{eq:unit_simplex}
\end{equation}
$f=(f_l)^k_{l=1} \in X^k$ is the discretized weighting function or the cost per pixel, $u=(u_l)^k_{l=1}$ is the primal variable and $X^k$ is the extension of the vector space for $k$ classes.
Considering 
\begin{equation}
\begin{split}
&\frac{1}{2} \sum_{l=1}^{k} \left(\int_{\Omega}|\nabla u_l| + \inner{u_l}{f_l} \right) = \\
&\max_{ p=(p_l)_{l=1}^k} \left( \sum_{l=1}^{k} \Big( \inner{\nabla u_l}{p_l} + \inner{u_l}{f_l} \Big) \right) - \delta_P(p),
\end{split}
\end{equation}
where $p \in Y^k$ is the dual variable, with $Y^k$ is the extension of the gradient vector space for $k$ classes, $\delta_P(p)$ is the indicator function for $p \in P$ defined as 
\begin{equation}
P = \left\{ p \in Y^k : ||p_l||_{\infty} \le \frac{1}{2} \right\}.
\label{eq:unit_balls}
\end{equation}
We have the primal-dual formulation as
\begin{equation}
\min_{ u=(u_l)_{l=1}^k} \max_{ p=(p_l)_{l=1}^k} \left( \sum_{l=1}^{k} \Big( \inner{\nabla u_l}{p_l} + \inner{u_l}{f_l} \Big) \right) + \delta_U(u) - \delta_P(p),
\label{eq:relaxation_primaldual}
\end{equation}
with $u,p$ related as $\inner{\nabla u}{p}_Y = -\inner{u}{\text{div} p}_X$. This result is a consequence of applying Guass Divergence theorem on the scalar function $u$ and vector field $p$. A corollary of the fore mentioned result is the relation $-\text{div}=\nabla^*$ where $\text{div}, \nabla^*$ are divergence in $Y$;  and the conjugate of gradient $\nabla$, respectively. Further since $\nabla^* = -\nabla^T$ it turns out that $\text{div}=\nabla^T$.\\

{\bf \emph{Bregman Proximity Functions:}}
The sets considered so far are unit simplex and unit ball (or more strictly a ball of radius $\frac{1}{2}$) as defined by $U,P$ in Eqs.\ref{eq:unit_simplex},\ref{eq:unit_balls}, respectively. The resolvent of these sets are orthogonal projections on unit simplex and point projection onto unit ball, respectively. However, in the case of segmentation when using more sophisticated relaxation that yield better delineation along edges, like \emph{paired calibrations} is used in Eq.\ref{eq:relaxation_seg} given by,
\begin{equation}
\begin{split}
& J(u) = \int_{\Omega} \Psi(Du); \\
s.t \hspace{2mm} &\Psi(a) = \sup_{b}\left\{ \sum_{l=1}^{k} \inner{a_l}{b_m} : |a_l - b_m| \le 1, 1\le l \le m \le k \right\},
\end{split}
\end{equation}
where $a=(a_1,\cdots,a_k), b=(b_1,\cdots,b_k)$. The corresponding set for the dual variables is no longer a unit ball, but intersection of unit balls given by,
\begin{equation}
P =\left\{ p \in Y^k : |p_l - p_m|_{\infty} \le 1, 1\le l \le m \le k \right\}.
\label{eq:inter_1balls}
\end{equation}
The resolvent of which, is an orthogonal projection on to such an intersection of unit balls. As the relaxations get more sophisticated the corresponding resolvent set becomes more complex and orthogonal projections on to them get computationally more involved. One approach towards getting a solution that can be used in a computational algorithm is to use \emph{Bregman functions}. Suppose we have a convex function $\psi(x)$ that is continuously differentiable on the interior of its domain; $\text{int}(X)$ and continuous on its closure; $\text{cl}(X)$, we use $\bar{x}$ to denote a point from $\text{int}(X)$. A Bregman proximity function $D_{\psi} : X \times \text{int}(X) \rightarrow \mathbb{R}$ generated by $\psi$ is defined as 
\begin{equation}
D_{\psi}(x,\bar{x}) = \psi(x)-\psi(\bar{x}) - \inner{\nabla \psi(\bar{x})}{x-\bar{x}}.
\label{eq:bregman}
\end{equation}

In iterative algorithms, the Bregman proximity function can be used with the proximity operator for a convex function $g : X \rightarrow \mathbb{R}$ as
\begin{equation}
\text{prox}_{\alpha g}^{\psi}(\bar{x}) = \arg \min_{x \in X} \alpha g(x) + D_{\psi}(x,\bar{x}).
\label{eq:bregman_prox}
\end{equation}

In image segmentation problem the basic class of functions of interest are of the form $g(x) = \inner{x}{c} + \delta_X(x)$ as seen in Eq.\ref{eq:relaxation_primaldual}. The associated proximal operator is 
\begin{equation}
\text{prox}_{\alpha g}^{\psi}(\bar{x}) = \arg \min_{x \in X} \alpha \inner{x}{c} + D_{\psi}(x,\bar{x}).
\label{prox_breg}
\end{equation}
The necessary and sufficient condition for optimality, which has a unique solution 
for Eq.\ref{prox_breg} is
\begin{equation}
\nabla \psi(\bar{x}) - c = \nabla \psi(x).
\end{equation}
This constraint is implicitly taken care by the Bregman proximity function. For further details on Bregman functions one may refer \cite{pdNet_Ochs}. In image segmentation the dual variables belong to the intersection of unit balls as shown in Eq.\ref{eq:inter_1balls} so each coordinate of the dual variable $p$ should satisfy $-1 \le p_j \le 1$ and solve the dual problem
\begin{equation}
\max_{ p=(p_l)_{l=1}^k} \sum_{l=1}^{k} \inner{\nabla u_l}{p_l}  - \delta_P(p).
\end{equation}
A suitable Bregman proximity functions that encode the dual variable constraints and the corresponding proximity solution along each coordinate $i$ are given by
\begin{equation}
\begin{split}
& \psi(x)= \frac{1}{2} \left[ (1+x)\log (1+x) + (1-x)\log (1-x) \right]; \\
& \left( \text{prox}_{\alpha g}^{\psi}(\bar{x}) \right)_i = \frac{\exp \left( -2 \alpha c_i \right) - \frac{1-\bar{x}_i}{1+\bar{x}_i}}{\exp \left( -2 \alpha c_i \right) + \frac{1-\bar{x}_i}{1+\bar{x}_i}},
\end{split}
\label{dual_solution}
\end{equation}
where $c_i = (\nabla u)_i$ is the $i$-component of the $\nabla u$. Similarly, the primal problem deals with 
\begin{equation}
\min_{ u=(u_l)_{l=1}^k} \inner{u_l}{f_l}  + \delta_U(u),
\end{equation}
with the primal variables $u$ restricted by $u_i \ge 0$ along each coordinate. The Bregman function encoding these constraints and the corresponding proximal solution are given by
\begin{equation}
\begin{split}
& \psi(x)= x \log x, \\
& \left( \text{prox}_{\alpha g}^{\psi}(\bar{x}) \right)_i = x_i \exp \left( -2 \alpha c_i \right).
\end{split}
\label{primal_solution_unnorm}
\end{equation}
To satisfy $\sum_{l=1}^{k} u_l(x)=1$ in Eq.\ref{eq:unit_simplex} it is sufficient to normalize it as 
\begin{equation}
 \left( \text{prox}_{\alpha g}^{\psi}(\bar{x}) \right)_i = \frac{x_i \exp \left( -2 \alpha c_i \right)}{ \sum_{j=1}^{K}x_j \exp \left( -2 \alpha c_j \right)},
\label{primal_solution}
\end{equation}
where $c_i = (\nabla^T p)_i - f_i$, $f_i$ denoting the cost associated with the $i^{th}$ class. Eqs.\ref{dual_solution} and \ref{primal_solution}, can now be used in Eq.\ref{eq:proximal_alog} to converge to a solution for image segmentation.

\end{document}